# Autoencoding beyond pixels using a learned similarity metric

**Anders Boesen Lindbo Larsen**[1]    ABLL@DTU.DK
**Søren Kaae Sønderby**[2]    SKAAESONDERBY@GMAIL.COM
**Hugo Larochelle**[3]    HLAROCHELLE@TWITTER.COM
**Ole Winther**[1,2]    OLWI@DTU.DK

[1] Department for Applied Mathematics and Computer Science, Technical University of Denmark
[2] Bioinformatics Centre, Department of Biology, University of Copenhagen, Denmark
[3] Twitter, Cambridge, MA, USA

## Abstract

We present an autoencoder that leverages learned representations to better measure similarities in data space. By combining a variational autoencoder with a generative adversarial network we can use learned feature representations in the GAN discriminator as basis for the VAE reconstruction objective. Thereby, we replace element-wise errors with feature-wise errors to better capture the data distribution while offering invariance towards e.g. translation. We apply our method to images of faces and show that it outperforms VAEs with element-wise similarity measures in terms of visual fidelity. Moreover, we show that the method learns an embedding in which high-level abstract visual features (e.g. wearing glasses) can be modified using simple arithmetic.

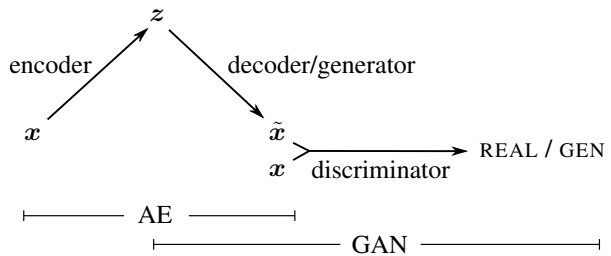

*Figure 1.* Overview of our network. We combine a VAE with a GAN by collapsing the decoder and the generator into one.

## 1. Introduction

Deep architectures have allowed a wide range of discriminative models to scale to large and diverse datasets. However, generative models still have problems with complex data distributions such as images and sound. In this work, we show that currently used similarity metrics impose a hurdle for learning good generative models and that we can improve a generative model by employing a learned similarity measure.

When learning models such as the variational autoencoder (VAE) (Kingma & Welling, 2014; Rezende et al., 2014), the choice of similarity metric is central as it provides the main part of the training signal via the reconstruction error objective. For this task, element-wise measures like the squared error are the default. Element-wise metrics are simple but not very suitable for image data, as they do not model the properties of human visual perception. E.g. a small image translation might result in a large pixel-wise error whereas a human would barely notice the change. Therefore, we argue in favor of measuring image similarity using a higher-level and *sufficiently invariant* representation of the images. Rather than hand-engineering a suitable measure to accommodate the problems of element-wise metrics, we want to learn a function for the task. The question is how to learn such a similarity measure? We find that by jointly training a VAE and a generative adversarial network (GAN) (Goodfellow et al., 2014) we can use the GAN discriminator to measure sample similarity. We achieve this by combining a VAE with a GAN as shown in Fig. 1. We collapse the VAE decoder and the GAN generator into one by letting them share parameters and training them jointly. For the VAE training objective, we replace the typical element-wise reconstruction metric with a feature-wise metric expressed in the discriminator.

### 1.1. Contributions

Our contributions are as follows:

- We combine VAEs and GANs into an unsupervised generative model that simultaneously learns to *encode*, *generate* and *compare* dataset samples.

- We show that generative models trained with learned similarity measures produce better image samples than models trained with element-wise error measures.

- We demonstrate that unsupervised training results in a latent image representation with disentangled factors of variation (Bengio et al., 2013). This is illustrated in experiments on a dataset of face images labelled with visual attribute vectors, where it is shown that simple arithmetic applied in the learned latent space produces images that reflect changes in these attributes.

## 2. Autoencoding with learned similarity

In this section we provide background on VAEs and GANs. Then, we introduce our method for combining both approaches, which we refer to as VAE/GAN. As we'll describe, our proposed hybrid is motivated as a way to improve VAE, so that it relies on a more meaningful, feature-wise metric for measuring reconstruction quality during training.

### 2.1. Variational autoencoder

A VAE consists of two networks that *encode* a data sample $x$ to a latent representation $z$ and *decode* the latent representation back to data space, respectively:

$$z \sim \text{Enc}(x) = q(z|x) , \quad \tilde{x} \sim \text{Dec}(z) = p(x|z) . \quad (1)$$

The VAE regularizes the encoder by imposing a prior over the latent distribution $p(z)$. Typically $z \sim \mathcal{N}(\mathbf{0}, \mathbf{I})$ is chosen. The VAE loss is minus the sum of the expected log likelihood (the reconstruction error) and a prior regularization term:

$$\mathcal{L}_{\text{VAE}} = -\mathbb{E}_{q(z|x)}\left[\log\frac{p(x|z)p(z)}{q(z|x)}\right] = \mathcal{L}_{\text{llike}}^{\text{pixel}} + \mathcal{L}_{\text{prior}} \quad (2)$$

with

$$\mathcal{L}_{\text{llike}}^{\text{pixel}} = -\mathbb{E}_{q(z|x)}\left[\log p(x|z)\right] \quad (3)$$
$$\mathcal{L}_{\text{prior}} = D_{\text{KL}}(q(z|x)\|p(z)) , \quad (4)$$

where $D_{\text{KL}}$ is the Kullback-Leibler divergence.

### 2.2. Generative adversarial network

A GAN consists of two networks: the *generator* network $\text{Gen}(z)$ maps latents $z$ to data space while the *discriminator* network assigns probability $y = \text{Dis}(x) \in [0, 1]$ that $x$ is an actual training sample and probability $1 - y$ that $x$ is generated by our model through $x = \text{Gen}(z)$ with $z \sim p(z)$. The GAN objective is to find the binary classifier that gives the best possible discrimination between true and generated data and simultaneously encouraging Gen to fit the true data distribution. We thus aim to maximize/minimize the binary cross entropy:

$$\mathcal{L}_{\text{GAN}} = \log(\text{Dis}(x)) + \log(1 - \text{Dis}(\text{Gen}(z))) , \quad (5)$$

with respect to Dis / Gen with $x$ being a training sample and $z \sim p(z)$.

### 2.3. Beyond element-wise reconstruction error with VAE/GAN

An appealing property of GAN is that its discriminator network implicitly has to learn a rich similarity metric for images, so as to discriminate them from "non-images". We thus propose to exploit this observation so as to transfer the properties of images learned by the discriminator into a more abstract reconstruction error for the VAE. The end result will be a method that combines the advantage of GAN as a high quality generative model and VAE as a method that produces an encoder of data into the latent space $z$.

Specifically, since element-wise reconstruction errors are not adequate for images and other signals with invariances, we propose replacing the VAE reconstruction (expected log likelihood) error term from Eq. 3 with a reconstruction error expressed in the GAN discriminator. To achieve this, let $\text{Dis}_l(x)$ denote the hidden representation of the $l$th layer of the discriminator. We introduce a Gaussian observation model for $\text{Dis}_l(x)$ with mean $\text{Dis}_l(\tilde{x})$ and identity covariance:

$$p(\text{Dis}_l(x)|z) = \mathcal{N}(\text{Dis}_l(x)|\text{Dis}_l(\tilde{x}), \mathbf{I}) , \quad (6)$$

where $\tilde{x} \sim \text{Dec}(z)$ is the sample from the decoder of $x$. We can now replace the VAE error of Eq. 3 with

$$\mathcal{L}_{\text{llike}}^{\text{Dis}_l} = -\mathbb{E}_{q(z|x)}[\log p(\text{Dis}_l(x)|z)] \quad (7)$$

We train our combined model with the triple criterion

$$\mathcal{L} = \mathcal{L}_{\text{prior}} + \mathcal{L}_{\text{llike}}^{\text{Dis}_l} + \mathcal{L}_{\text{GAN}} . \quad (8)$$

Notably, we optimize the VAE wrt. $\mathcal{L}_{\text{GAN}}$ which we regard as a *style* error in addition to the reconstruction error which can be interpreted as a *content* error using the terminology from Gatys et al. (2015). Moreover, since both Dec and Gen map from $z$ to $x$, we share the parameters between the two (or in other words, we use Dec instead of Gen in Eq. 5).

In practice, we have observed the devil in the details during development and training of this model. We therefore provide a list of practical considerations in this section. We refer to Fig. 2 and Alg. 1 for overviews of the training procedure.

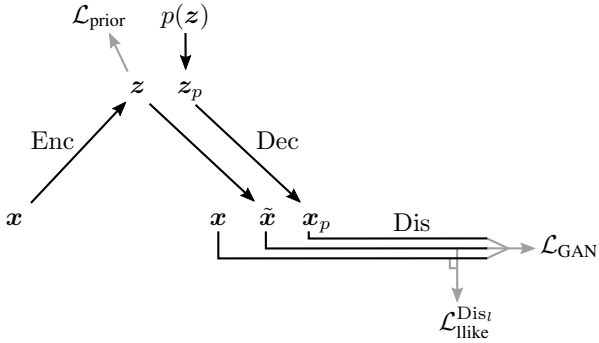

*Figure 2.* Flow through the combined VAE/GAN model during training. Gray lines represent terms in the training objective.

**Algorithm 1** Training the VAE/GAN model

$\boldsymbol{\theta}_{\text{Enc}}, \boldsymbol{\theta}_{\text{Dec}}, \boldsymbol{\theta}_{\text{Dis}} \leftarrow$ initialize network parameters
**repeat**
    $\boldsymbol{X} \leftarrow$ random mini-batch from dataset
    $\boldsymbol{Z} \leftarrow \text{Enc}(\boldsymbol{X})$
    $\mathcal{L}_{\text{prior}} \leftarrow D_{\text{KL}}(q(\boldsymbol{Z}|\boldsymbol{X})\|p(\boldsymbol{Z}))$
    $\tilde{\boldsymbol{X}} \leftarrow \text{Dec}(\boldsymbol{Z})$
    $\mathcal{L}_{\text{llike}}^{\text{Dis}_l} \leftarrow -\mathbb{E}_{q(\boldsymbol{Z}|\boldsymbol{X})}[p(\text{Dis}_l(\boldsymbol{X})|\boldsymbol{Z})]$
    $\boldsymbol{Z}_p \leftarrow$ samples from prior $\mathcal{N}(\boldsymbol{0}, \boldsymbol{I})$
    $\boldsymbol{X}_p \leftarrow \text{Dec}(\boldsymbol{Z}_p)$
    $\mathcal{L}_{\text{GAN}} \leftarrow \log(\text{Dis}(\boldsymbol{X})) + \log(1 - \text{Dis}(\tilde{\boldsymbol{X}}))$
        $+ \log(1 - \text{Dis}(\boldsymbol{X}_p))$
    // Update parameters according to gradients
    $\boldsymbol{\theta}_{\text{Enc}} \xleftarrow{+} -\nabla_{\boldsymbol{\theta}_{\text{Enc}}}(\mathcal{L}_{\text{prior}} + \mathcal{L}_{\text{llike}}^{\text{Dis}_l})$
    $\boldsymbol{\theta}_{\text{Dec}} \xleftarrow{+} -\nabla_{\boldsymbol{\theta}_{\text{Dec}}}(\gamma \mathcal{L}_{\text{llike}}^{\text{Dis}_l} - \mathcal{L}_{\text{GAN}})$
    $\boldsymbol{\theta}_{\text{Dis}} \xleftarrow{+} -\nabla_{\boldsymbol{\theta}_{\text{Dis}}} \mathcal{L}_{\text{GAN}}$
**until** deadline

**Limiting error signals to relevant networks** Using the loss function in Eq. 8, we train both a VAE and a GAN simultaneously. This is possible because we do not update all network parameters wrt. the combined loss. In particular, Dis should not try to minimize $\mathcal{L}_{\text{llike}}^{\text{Dis}_l}$ as this would collapse the discriminator to 0. We also observe better results by not backpropagating the error signal from $\mathcal{L}_{\text{GAN}}$ to Enc.

**Weighting VAE vs. GAN** As Dec receives an error signal from both $\mathcal{L}_{\text{llike}}^{\text{Dis}_l}$ and $\mathcal{L}_{\text{GAN}}$, we use a parameter $\gamma$ to weight the ability to reconstruct vs. fooling the discriminator. This can also be interpreted as weighting *style* and *content*. Rather than applying $\gamma$ to the entire model (Eq. 8), we perform the weighting only when updating the parameters of Dec:

$$\boldsymbol{\theta}_{\text{Dec}} \xleftarrow{+} -\nabla_{\boldsymbol{\theta}_{\text{Dec}}}(\gamma \mathcal{L}_{\text{llike}}^{\text{Dis}_l} - \mathcal{L}_{\text{GAN}}) \qquad (9)$$

**Discriminating based on samples from $p(\boldsymbol{z})$ and $q(\boldsymbol{z}|\boldsymbol{x})$**
We observe better results when using samples from $q(\boldsymbol{z}|\boldsymbol{x})$ (i.e. the encoder Enc) in addition to our prior $p(\boldsymbol{z})$ in the GAN objective:

$$\begin{aligned}\mathcal{L}_{\text{GAN}} = \;& \log(\text{Dis}(\boldsymbol{x})) + \log(1 - \text{Dis}(\text{Dec}(\boldsymbol{z}))) \\ & + \log(1 - \text{Dis}(\text{Dec}(\text{Enc}(\boldsymbol{x}))))\end{aligned} \qquad (10)$$

Note that the regularization of the latent space $\mathcal{L}_{\text{prior}}$ should make the set of samples from either $p(\boldsymbol{z})$ or $q(\boldsymbol{z}|\boldsymbol{x})$ similar. However, for any given example $\boldsymbol{x}$, the negative sample $\text{Dec}(\text{Enc}(\boldsymbol{x}))$ is much more likely to be similar to $\boldsymbol{x}$ than $\text{Dec}(\boldsymbol{z})$. When updating according to $\mathcal{L}_{\text{GAN}}$, we suspect that having similar positive and negative samples makes for a more useful learning signal.

## 3. Related work

Element-wise distance measures are notoriously inadequate for complex data distributions like images. In the computer vision community, preprocessing images is a prevalent solution to improve robustness to certain perturbations. Examples of preprocessing are contrast normalization, working with gradient images or pixel statistics gathered in histograms. We view these operations as a form of metric engineering to account for the shortcomings of simple element-wise distance measures. A more detailed discussion on the subject is provided by Wang & Bovik (2009).

Neural networks have been applied to metric learning in form of the Siamese architecture (Bromley et al., 1993; Chopra et al., 2005). The learned distance metric is minimized for similar samples and maximized for dissimilar samples using a max margin cost. However, since Siamese networks are trained in a supervised setup, we cannot apply them directly to our problem.

Several attempts at improving on element-wise distances for generative models have been proposed within the last year. Ridgeway et al. (2015) apply the structural similarity index as an autoencoder (AE) reconstruction metric for grey-scale images. Yan et al. (2015) let a VAE output two additional images to learn shape and edge structures more explicitly. Mansimov et al. (2015) append a GAN-based sharpening step to their generative model. Mathieu et al. (2015) supplement a squared error measure with both a GAN and an image gradient-based similarity measure to improve image sharpness of video prediction. While all these extensions yield visibly sharper images, they do not have the same potential for capturing high-level structure compared to a deep learning approach.

In contrast to AEs that model the relationship between a dataset sample and a latent representation directly, GANs learn to generate samples indirectly. By optimizing the GAN generator to produce samples that imitate the dataset according to the GAN discriminator, GANs avoid element-wise similarity measures by construction. This is a likely explanation for their ability to produce high-quality images as demonstrated by Denton et al. (2015); Radford et al. (2015).

Lately, convolutional networks with upsampling have shown useful for generating images from a latent representation. This has sparked interest in learning image embeddings where semantic relationships can be expressed using simple arithmetic – similar to the suprising results of the *word2vec* model by Mikolov et al. (2013). First, Dosovitskiy et al. (2015) used supervised training to train convolutional network to generate chairs given high-level information about the desired chair. Later, Kulkarni et al. (2015); Yan et al. (2015); Reed et al. (2015) have demonstrated encoder-decoder architectures with disentangled feature representations, but their training schemes rely on supervised information. Radford et al. (2015) inspect the latent space of a GAN after training and find directions corresponding to eyeglasses and smiles. As they rely on pure GANs, however, they cannot encode images making it challenging to explore the latent space.

Our idea of a learned similarity metric is partly motivated by the neural artistic style network of Gatys et al. (2015) who demonstrate the representational power of deep convolutional features. They obtain impressive results by optimizing an image to have similar features as a subject image and similar feature correlations as a style image in a pretrained convolutional network. In our VAE/GAN model, one could view $\mathcal{L}_{\text{llike}}^{\text{Dis}_l}$ as content and $\mathcal{L}_{\text{GAN}}$ as style. Our style term, though, is not computed from feature correlations but is the error signal from trying to fool the GAN discriminator.

## 4. Experiments

Measuring the quality of generative models is challenging as current evaluation methods are problematic for larger natural images (Theis et al., 2015). In this work, we use images of size 64x64 and focus on more qualitative assessments since traditional log likelihood measures do not capture visual fidelity. Indeed, we have tried discarding the GAN discriminator after training of the VAE/GAN model and computing a pixel-based log likelihood using the remaining VAE. The results are far from competitive with plain VAE models (on the CIFAR-10 dataset).

In this section we investigate the performance of different generative models:

- Plain VAE with an element-wise Gaussian observation model.

- VAE with a learned distance ($\text{VAE}_{\text{Dis}_l}$). We first train a GAN and use the discriminator network as a learned similarity measure. We select a single layer $l$ at which we measure the similariy according to $\text{Dis}_l$. $l$ is chosen such that the comparison is performed after 3 downsamplings of each a factor of 2 in the convolutional encoder.
- The combined VAE/GAN model. This model is similar to $\text{VAE}_{\text{Dis}_l}$ but we also optimize Dec wrt. $\mathcal{L}_{\text{GAN}}$.
- GAN. This modes has recently been shown capable of generating high-quality images (Radford et al., 2015).

All models share the same architectures for Enc, Dec and Dis respectively. For all our experiments, we use convolutional architectures and use *backward convolution* (aka. *fractional striding*) with stride 2 to upscale images in Dec. Backward convolution is achieved by flipping the convolution direction such that striding causes upsampling. Our models are trained with RMSProp using a learning rate of 0.0003 and a batch size of 64. In table 1 we list the network architectures. We refer to our implementation available online[1].

### 4.1. CelebA face images

We apply our methods to face images from the *CelebA* dataset[2] (Liu et al., 2015). This dataset consists of 202,599 images annotated with 40 binary attributes such as *eyeglasses*, *bangs*, *pale skin* etc. We scale and crop the images to 64×64 pixels and use only the images (not the attributes) for unsupervised training.

After training, we draw samples from $p(z)$ and propagate these through Dec to generate new images which are shown in Fig. 3. The plain VAE is able draw the frontal part of the face sharply, but off-center the images get blurry. This is because the dataset aligns faces using frontal landmarks. When we move too far away from the aligned parts, the recognition model breaks down because pixel correspondence cannot be assumed. $\text{VAE}_{\text{Dis}_l}$ produces sharper images even off-center because the reconstruction error is lifted beyond pixels. However, we see severe noisy artefacts which we believe are caused by the harsh downsampling scheme. In comparison, VAE/GAN and pure GAN produce sharper images with more natural textures and face parts.

Additionally, we make the VAEs reconstruct images taken from a separate test set. Reconstruction is not possible with the GAN model as it lacks an encoder network. The results are shown in Fig. 4 and our conclusions are similar to what

---

[1] http://github.com/andersbll/autoencoding_beyond_pixels

[2] We use the aligned and cropped version of the dataset.

**Autoencoding beyond pixels using a learned similarity metric**

| Enc | Dec | Dis |
|---|---|---|
| 5×5 64 conv. ↓, BNorm, ReLU | 8·8·256 fully-connected, BNorm, ReLU | 5×5 32 conv., ReLU |
| 5×5 128 conv. ↓, BNorm, ReLU | 5×5 256 conv. ↑, BNorm, ReLU | 5×5 128 conv. ↓, BNorm, ReLU |
| 5×5 256 conv. ↓, BNorm, ReLU | 5×5 128 conv. ↑, BNorm, ReLU | 5×5 256 conv. ↓, BNorm, ReLU |
| 2048 fully-connected, BNorm, ReLU | 5×5 32 conv. ↑, BNorm, ReLU | 5×5 256 conv. ↓, BNorm, ReLU |
| | 5×5 3 conv., tanh | 512 fully-connected, BNorm, ReLU |
| | | 1 fully-connected, sigmoid |

*Table 1.* Architectures for the three networks that comprise VAE/GAN. ↓ and ↑ represent down- and upsampling respectively. BNorm denotes batch normalization (Ioffe & Szegedy, 2015). When batch normalization is applied to convolutional layers, per-channel normalization is used.

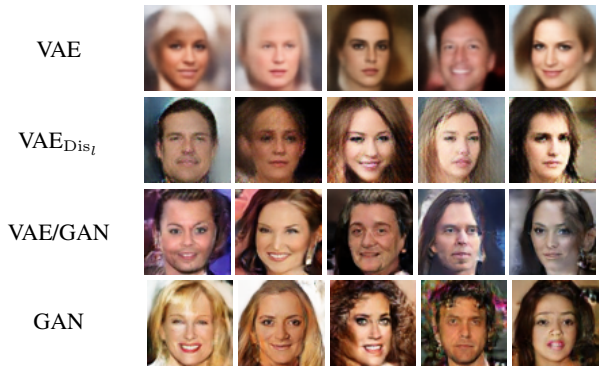

*Figure 3.* Samples from different generative models.

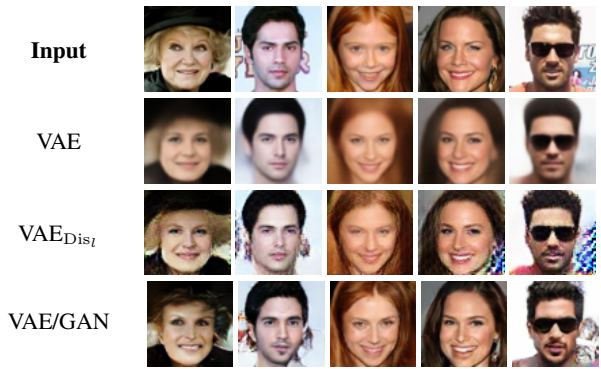

*Figure 4.* Reconstructions from different autoencoders.

we observed for the random samples. Note that VAE$_{Dis_l}$ generates noisy blue patterns in some of the reconstructions. We suspect the GAN-based similarity measure can collapse to 0 in certain cases (such as the pattern we observe), which encourages Dec to generate such patterns.

#### 4.1.1. VISUAL ATTRIBUTE VECTORS

Inspired by attempts at learning embeddings in which semantic concepts can be expressed using simple arithmetic (Mikolov et al., 2013), we inspect the latent space of a trained VAE/GAN model. The idea is to find directions in the latent space corresponding to specific visual features in image space.

We use the binary attributes of the dataset to extract *visual attribute vectors*. For all images we use the encoder to calculate latent vector representations. For each attribute, we compute the mean vector for images with the attribute and the mean vector for images without the attribute. We then compute the visual attribute vector as the difference between the two mean vectors. This is a very simple method for calculating visual attribute vectors that will have problems with highly correlated visual attributes such as *heavy makeup* and *wearing lipstick*. In Fig. 5, we show face images as well as the reconstructions after adding different visual attribute vectors to the latent representations. Though not perfect, we clearly see that the attribute vectors capture semantic concepts like *eyeglasses*, *bangs*, etc. E.g. when bangs are added to the faces, both the hair color and the hair texture matches the original face. We also see that being a man is highly correlated with having a mustache, which is caused by attribute correlations in the dataset.

### 4.2. Attribute similarity, Labeled faces in the wild

Inspired by the *attribute similarity* experiment of Yan et al. (2015), we seek a more quantative evaluation of our generated images. The idea is to learn a generative model for face images conditioned on facial attributes. At test time, we generate face images by retrieval from chosen attribute configurations and let a separately trained regressor network predict the attributes from the generated images. A good generative model should be able to produce visual attributes that are correctly recognized by the regression model. To imitate the original experiment, we use Labeled faces in the wild (LFW) images (Huang et al., 2007) with attributes (Kumar et al., 2009). We align the face images according to the landmarks in (Zhu et al., 2014). Additionally, we crop and resize the images to 64×64 pixels and augment the dataset with common operations. Again, we refer to our implementation online for more details.

We construct conditional VAE, GAN and VAE/GAN models by concatenating the attribute vector to the vector repre-sentation of the input in Enc, Dec and Dis similar to (Mirza & Osindero, 2014). For Enc and Dis, the attribute vector is concatenated to the input of the top fully connected layer. Our regression network has almost the same architecture as Enc. We train using the LFW training set, and during testing, we condition on the test set attributes and sample faces to be propagated through the regression network. Figure 6 shows faces generated by conditioning on attribute vectors from the test set. We report regressor performance numbers in Table 2. Compared to an ordinary VAE, the VAE/GAN model yields significantly better attributes visually that leads to smaller recognition error. The GAN network performs suprisingly poorly and we suspect that this is caused by instabilities during training (GAN models are very difficult to train reliably due to the minimax objective function). Note that our results are not directly comparable with those of Yan et al. (2015) since we do not have access to their preprocessing scheme nor regression model.

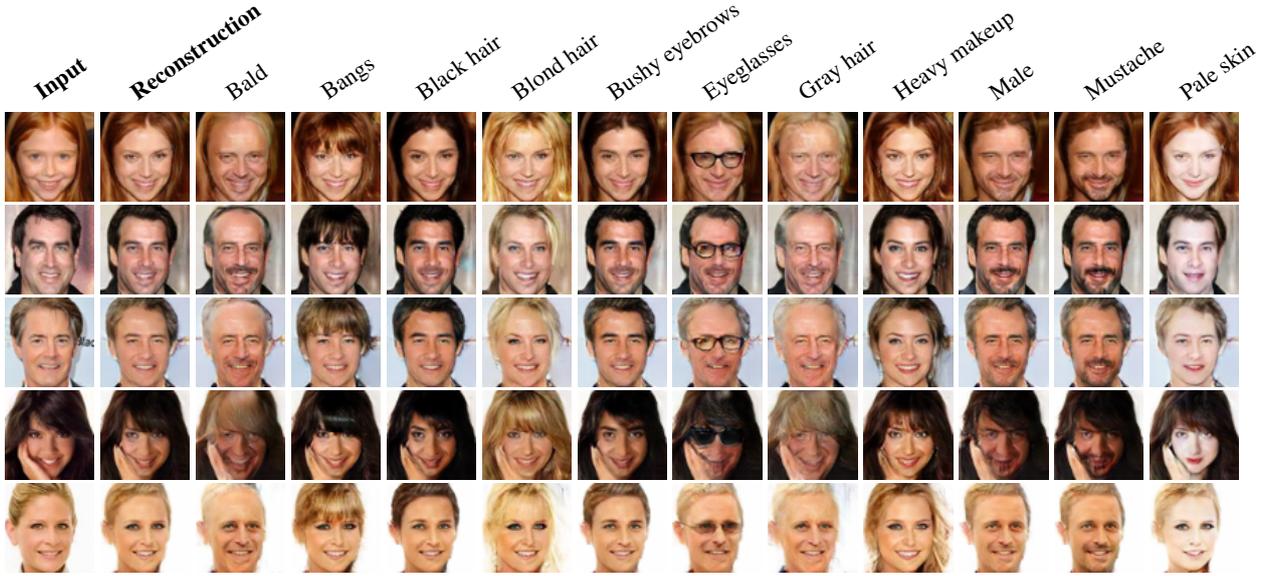

Figure 5. Using the VAE/GAN model to reconstruct dataset samples with visual attribute vectors added to their latent representations.

| Model | Cosine similarity | Mean squared error |
| --- | --- | --- |
| LFW test set | 0.9193 | 14.1987 |
| VAE | 0.9030 | 27.59 ± 1.42 |
| GAN | 0.8892 | 27.89 ± 3.07 |
| VAE/GAN | **0.9114** | **22.39 ± 1.16** |

Table 2. Attribute similarity scores. To replicate (Yan et al., 2015), the cosine similarity is measured as the best out of 10 samples per attribute vector from the test set. The mean squared error is computed over the test set and statistics are measured over 25 runs.

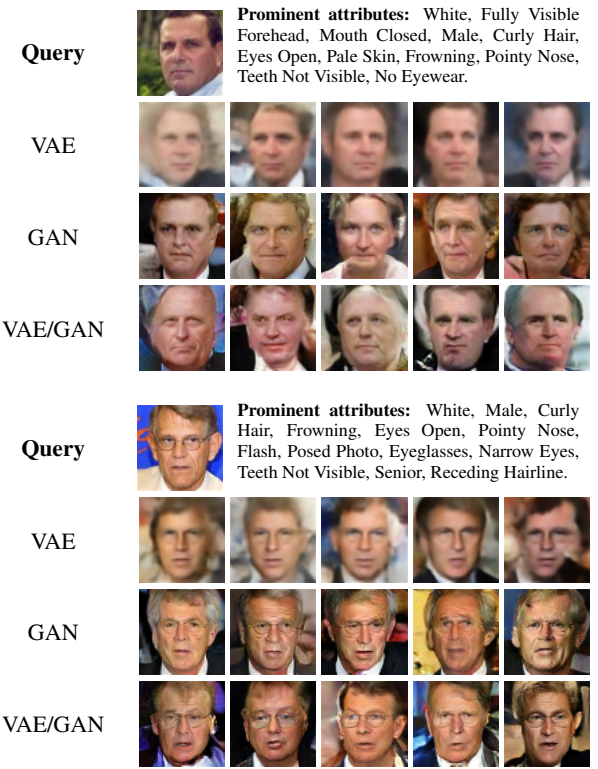

Figure 6. Generating samples conditioned on the LFW attributes listed alongside their corresponding image.

### 4.3. Unsupervised pretraining for supervised tasks

For completeness, we report that we have tried evaluating VAE/GAN in a semi-supervised setup by unsupervised pretraining followed by finetuning using a small number of labeled examples (for both CIFAR-10 and STL-10 datasets). Unfortunately, we have not been able to reach results competitive with the state-of-the-art (Rasmus et al., 2015; Zhao et al., 2015). We speculate that the intra-class variation may be too high for the VAE-GAN model to learn good generalizations of the different object classes.

## 5. Discussion

The problems with element-wise distance metrics are well known in the literature and many attempts have been made at going beyond pixels – typically using hand-engineered measures. Much in the spirit of deep learning, we argue that the similarity measure is yet another component which can be replaced by a learned model capable of capturing high-level structure relevant to the data distribution. In this work, our main contribution is an unsupervised scheme for learning and applying such a distance measure. With the learned distance measure we are able to train an image encoder-decoder network generating images of unprecedented visual fidelity as shown by our experiments. Moreover, we show that our network is able to disentangle factors of variation in the input data distribution and discover visual attributes in the high-level representation of the latent space. In principle, this lets us employ a large set of unlabeled images for training and use a small set of labeled images to discover features in latent space.

We regard our method as an extension of the VAE framework. Though, it must be noted that the high quality of our generated images is due to the combined training of Dec as a both a VAE decoder and a GAN generator. This makes our method more of a hybrid between VAE and GAN, and alternatively, one could view our method more as an extension of GAN where $p(\boldsymbol{z})$ is constrained by an additional network.

It is not obvious that the discriminator network of a GAN provides a useful similarity measure as it is trained for a different task, namely being able to tell generated samples from real samples. However, convolutional features are often surprisingly good for transfer learning, and as we show, good enough in our case to improve on element-wise distances for images. It would be interesting to see if better features in the distance measure would improve the model, e.g. by employing a similarity measure provided by a Siamese network trained on faces, though in practice Siamese networks are not a good fit with our method as they require labeled data. Alternatively one could investigate the effect of using a pretrained feedforward network for measuring similarity.

In summary, we have demonstrated a first attempt at unsupervised learning of encoder-decoder models as well as a similarity measure. Our results show that the visual fidelity of our method is competitive with GAN, which in that regard is considered state-of-the art. We therefore consider learned similarity measures a promising step towards scaling up generative models to more complex data distributions.

## Acknowledgements

We would like to thank Søren Hauberg, Casper Kaae Sønderby and Lars Maaløe for insightful discussions, Nvidia for donating GPUs used in experiments, and the authors of DeepPy[3] and CUDArray (Larsen, 2014) for the software frameworks used to implement our model.

---

[3] http://github.com/andersbll/deeppy